\documentclass[twoside,11pt]{article}

\usepackage{amssymb,latexsym,amsfonts,amsmath,url,epsf,graphicx}

\newcommand{\Ceq}{\stackrel{+}{=}}

\newcommand{\cP}{\cal P}

\newcommand{\R}{\mathbb R}

\newcommand\independent{\protect\mathpalette{\protect\independenT}{\perp}}
\def\independenT#1#2{\mathrel{\rlap{$#1#2$}\mkern2mu{#1#2}}}

\newtheorem{Postulate}{Postulate}
\newtheorem{Theorem}{Theorem}
\newtheorem{Definition}{Definition}
\newtheorem{Lemma}{Lemma}

\begin{document}

\title{Distinguishing Cause and Effect via
Second Order Exponential Models}

\author{Dominik Janzing\footnote{email: dominik.janzing@tuebingen.mpg.de}, Xiaohai Sun, and Bernhard Sch\"olkopf \\
{\small Max Planck Institute for Biological Cybernetics}\\
       {\small  72076 T{\"u}bingen, Germany}}

\date{29 October 2009}

\maketitle

\begin{abstract}
We propose a method to infer causal structures containing both discrete  and continuous  variables. 
The idea is to select causal hypotheses for  which the conditional density of every variable,  given its causes, becomes
smooth. We define a family of smooth densities and conditional densities by second order exponential models, i.e.,
by maximizing conditional entropy subject to first and  second statistical moments.
If some of the variables take only values in proper subsets of $\R^n$, these conditionals can induce  
different families of joint distributions even for Markov-equivalent graphs. 

We consider the  case of one binary and one real-valued variable where the method can distinguish between cause and effect.
Using this example, we describe that sometimes a causal hypothesis must be rejected because 
$P({\tt effect}|{\tt cause})$ and $P({\tt cause})$ 
share algorithmic information (which is untypical if they are chosen independently).
This way, our method is in the same spirit as faithfulness-based causal inference because it also rejects
non-generic mutual adjustments among DAG-parameters.  
\end{abstract}

\section{Introduction}

Finding causal structures that generated the statistical dependences among observed variables
has  attracted increasing interest in machine learning. 
Although there is in principle no method for reliably identifying causal structures
if no randomized studies are feasible, the seminal work of Spirtes et  al.~\cite{Spirtes1993} and Pearl \cite{Pearl2000} made it clear that under reasonable assumptions it is possible to derive causal information from purely observational data.

The formal language of the conventional approaches is a graphical model, where the random variables are the nodes of a directed acyclic graph (DAG) and an arrow from variable $X$ to $Y$ indicates that there is a direct causal influence from $X$ to $Y$. The definition of ``direct causal effect'' from $X$ to $Y$ refers to a hypothetical intervention where all variables in the model except from $X$ and $Y$ are adjusted to fixed values and one observes whether the distribution of $Y$ changes while $X$ is adjusted to different values. As clarified in  detail in \cite{Pearl2000}, the change of the distribution of $Y$ in such an intervention can be derived from the joint distribution of all relevant variables after the causal DAG is given. 

The essential postulate that connects statistics to causality is the so-called causal Markov condition stating 
that every variable is conditionally independent of its non-effects, given its  direct  causes \cite{Pearl2000}.
If the joint distribution of $X_1,\dots,X_n$ has a density $p(x_1,\dots,x_n)$ with respect to some product measure  
$\mu$ (which we assume throughout the paper),
the latter factorizes \cite{Lauritzen} into
\begin{equation}\label{Fac}
p(x_1,\dots,x_n)=\Pi_{j=1}^n p(x_j|pa_j)\,,
\end{equation}
where $pa_j$ is the set of all values of the parents of $X_j$ with respect  to the true causal graph.
 The conditional densities $p(x_j|pa_j)$  
will be called {\it Markov kernels}. They represent the mechanism that  generate the statistical dependences. 
 
A large class of known causal inference algorithms (like, for instance,  PC, IC, FCI, see \cite{Spirtes,Pearl2000}) are based
on the causal faithfulness principle which reads:
among all graphs that render the joint distribution Markovian, prefer those structures that allow {\it only} the observed conditional dependences. 
In other words, faithfulness is based on the assumption that all the observed independences are 
due to the causal structure rather than being a result of specific adjustments of parameters.
One of the main limitations of this type of independence-based causal  inference  is that 
there are typically a large number of DAGs that induce the same set  of independences.
Rules for the
selection of hypotheses within these {\it Markov  equivalence classes} are therefore desirable.

Before we describe our method, we briefly sketch some methods from the literature. 
\cite{Kano2003} have observed that linear causal relationships between
non-Gaussian distributed random variables induce joint measures which
require non-linear cause-effect relations for the wrong causal directions. Their causal
inference principle~\cite{Shimizu2005} for linear non-Gaussian acyclic
models (short: LiNGAM), is based on independent component
analysis.\footnote{It was implemented in Matlab by P. O. Hoyer, available at\\
http://www.cs.helsinki.fi/group/neuroinf/lingam/.} It selects causal
hypotheses for which almost linear cause-effect relations are
sufficient whenever such hypotheses are possible for a given
distribution.\footnote{Apart from this, it has  been shown that the linearity assumption helps also for causal inference in the  presence 
of latent variables (see, e.g. \cite{Silva2006}).}
\cite{Hoyer} generalized this idea to the case where every variable is a possibly (non-linear) function
of its  direct causes up to some additive  noise term that is independent of the causes (see also \cite{ICMLnonlingam} and  \cite{UAI_CAN}). Under this assumption, different causal structures induce, in the generic case,  different classes of  joint distributions 
even if the  causal graphs belong to the same equivalence class.  
\cite{Zhang,Zhang_UAI} generalized the model class to the case where every function is additionally subjected to non-linear distortion compared  to the models of \cite{Hoyer}.
However, all these algorithms  only work for
real-valued variables and the generalization to discrete variables is not straightforward. 

Here we describe a method (first proposed in our conference paper \cite{SunLauderdale}) 
that can deal with combinations of discrete and continuous variables, it even  benefits
from such a combination. More precisely, it requires that  at
  least one of the variables
is discrete or attains only values in a proper subset of $\R^n$.
We define a parametric family of conditionals that  induce 
different families of joint distributions for different causal directions. 
The underlying idea is related to an observation of \cite{Comley} stating 
that the same joint distribution of combinations  of discrete and continuous variables may have 
descriptions in terms of simple Markov kernels for one DAG but require more complex ones for other DAGs. 
In contrast  to \cite{Comley}, we define families  of Markov kernels that  are derived from a unique
principle, regardless of whether the variables are discrete or continuous.  

\begin{figure}
\centerline{\includegraphics[scale=0.40]{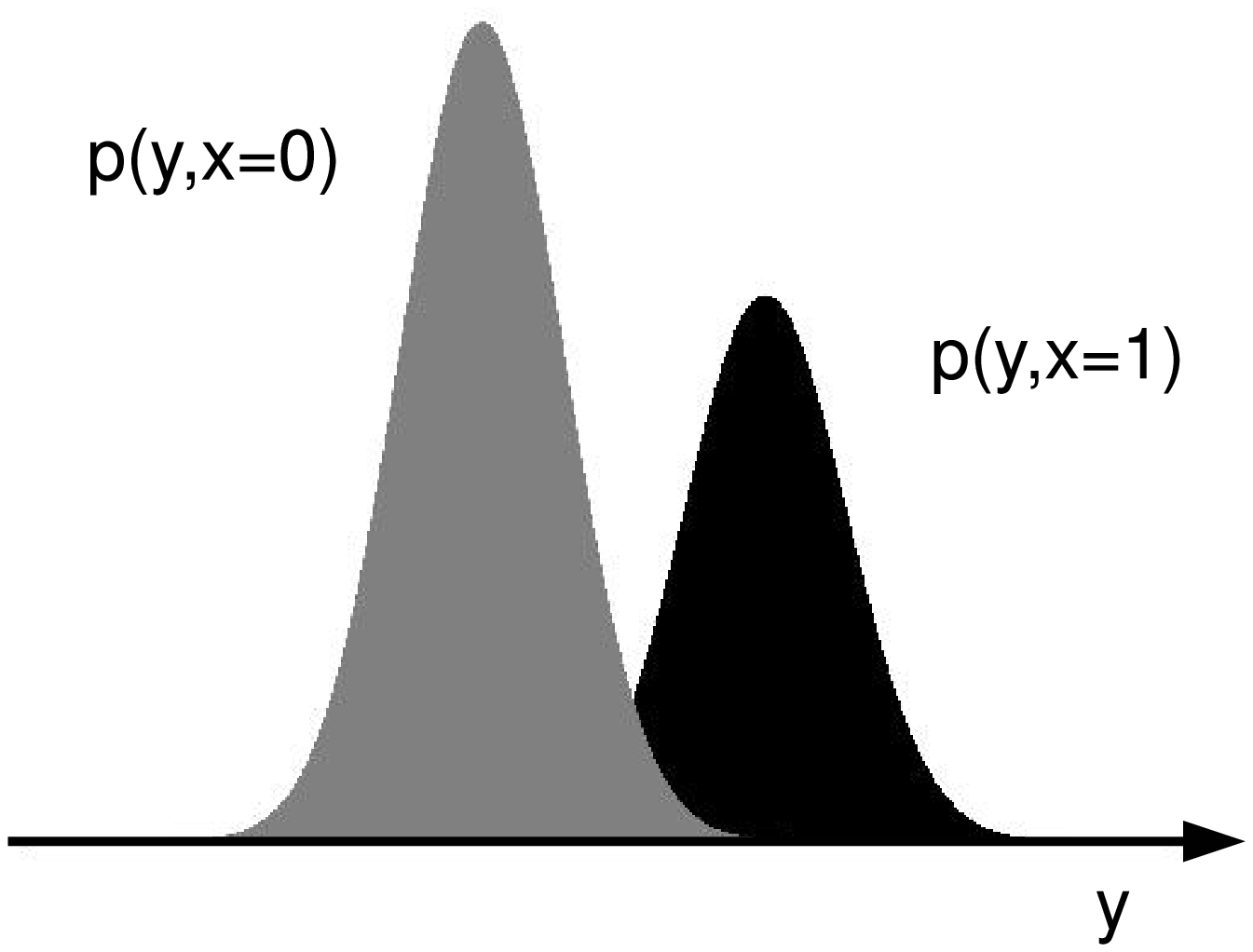}\hspace{-3cm}\includegraphics[scale=0.40]{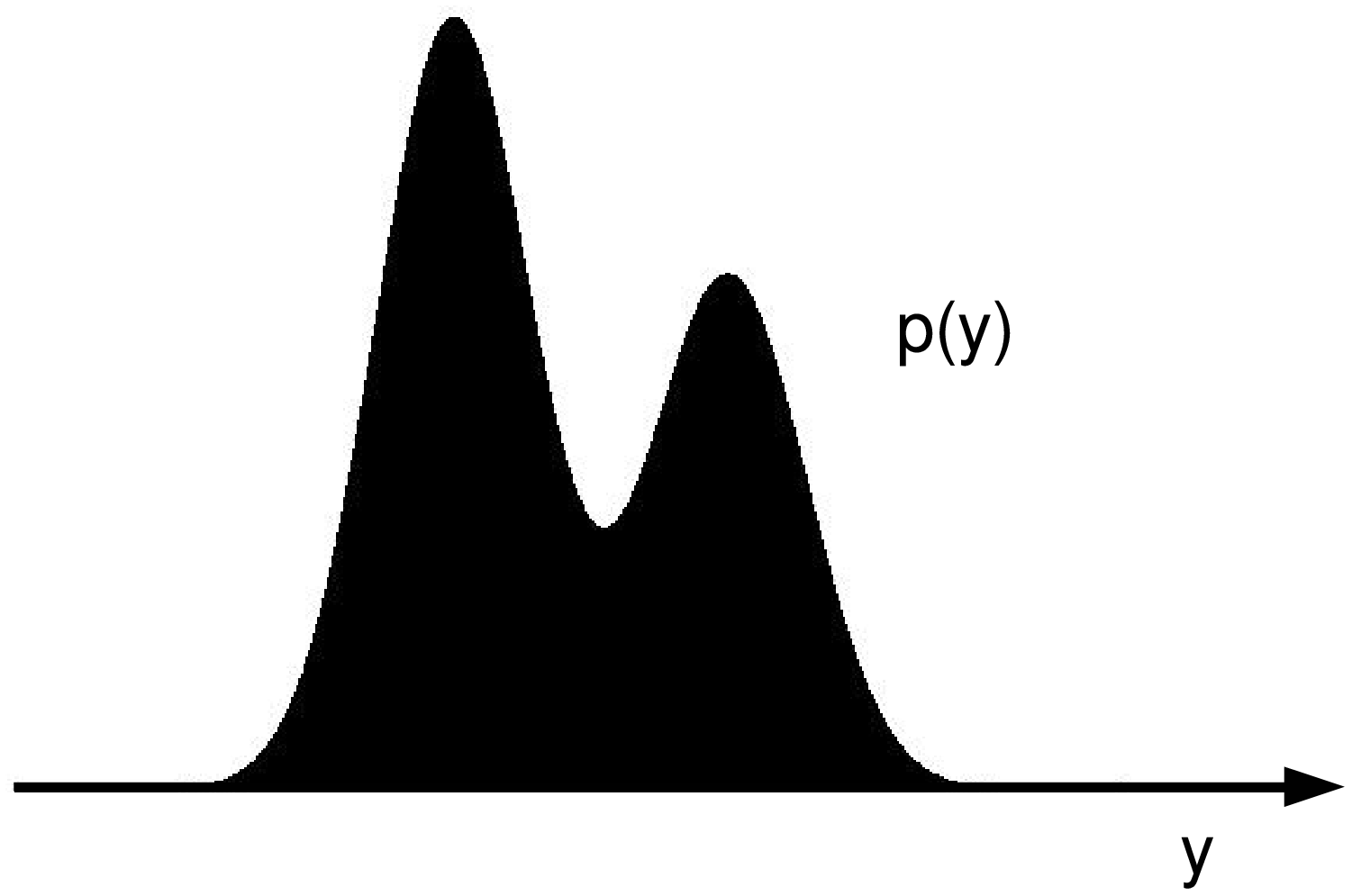}}
\caption{\label{bimodal}{\small Left: Joint density $p(x,y)$ of a real-valued variable $X$ and a binary variable $X$ suggesting a model $X\rightarrow Y$, because the influence of $X$ on $Y$ then only consists of shifting the mean of the gaussian. 
The causal hypothesis  
$Y\rightarrow X$ is less likely: only specific choices of $p(x|y)$ would separate the bimodal Gaussian mixture $p(y)$ (right) into two 
separate modes. It requires an even more specific conditional $p(x|y)$ to make the components 
gaussian.}}
\end{figure}

To describe our idea, assume that $X$ is  a binary variable and $Y$ real-valued and that we observe the joint distribution shown in
Fig.~\ref{bimodal}: Let  $p(y)$ be a bimodal mixture of  two Gaussians such that both $p(y|x=0)$ and $p(y|x=1)$ are Gaussians with the same width but different mean.
Then it is natural to assume that $X$ is the cause and $Y$ the  effect because changing the value of $X$ then would simply shift the mean of  $Y$.
For the converse model $Y\rightarrow  X$, bimodality of $Y$ remains unexplained. Moreover, it seems unlikely, that conditioning on  the effect $X$ separates the two modes of $p(y)$ even though 
$X$ is not causally responsible for the bimodality.  

\begin{figure}
\centerline{\includegraphics[scale=0.30]{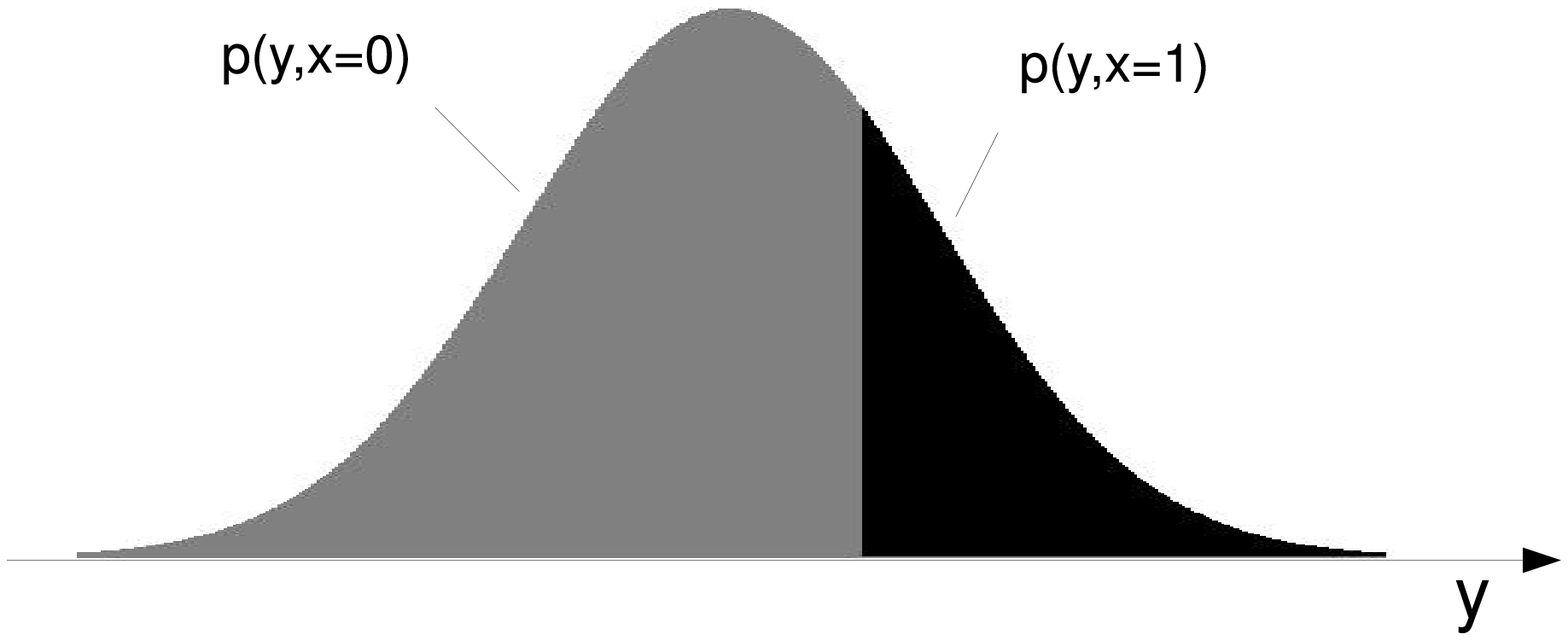}\hspace{0.5cm}\includegraphics[scale=0.3]{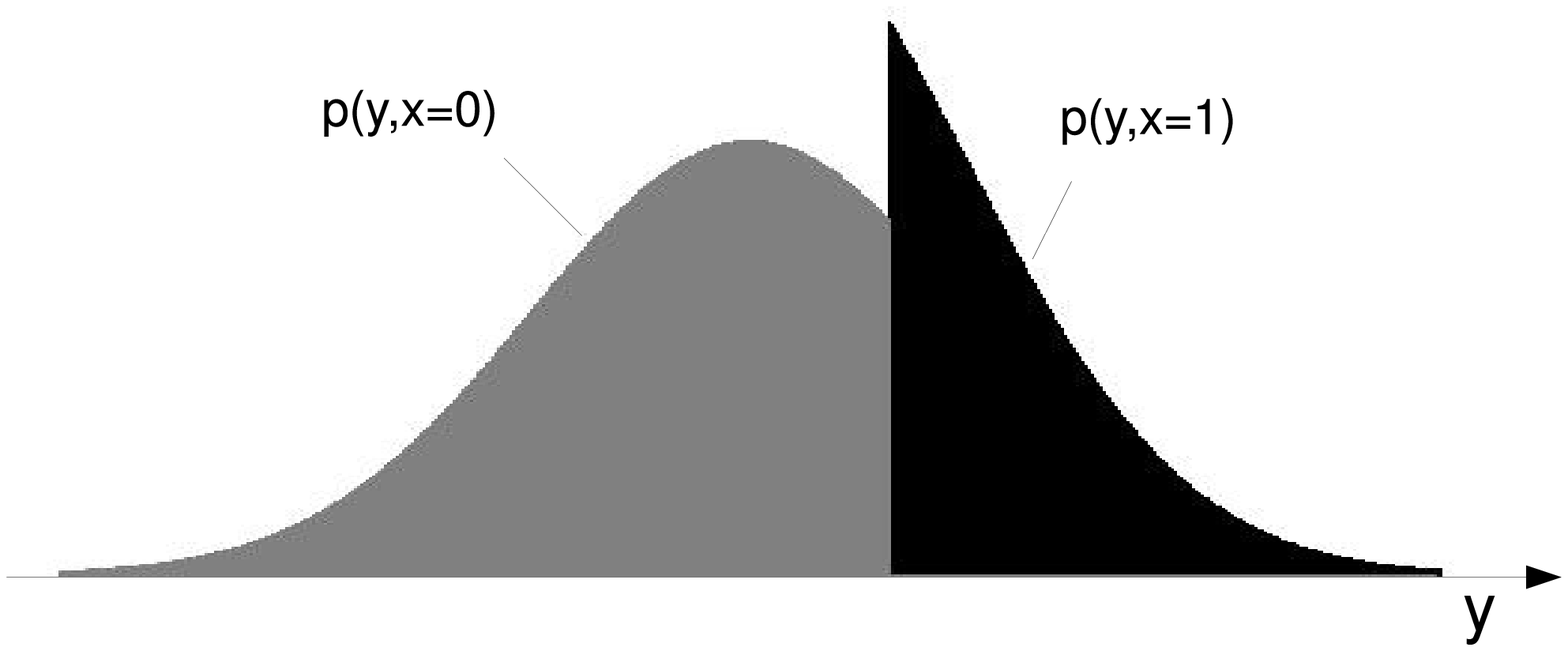}}
\caption{\label{Threshold}{\small Joint  density $p(x,y)$ of a real-valued random variable $Y$ and a binary variable  $X$. The marginal distribution $p(y)$  is Gaussian. The causal hypothesis $Y\rightarrow X$ is plausible: the conditional $p(x|y)$ 
corresponds to setting $x=1$ for all $y$ above a certain  threshold.
We reject the converse hypothesis $X\rightarrow Y$ 
because $p(y|x)$ and $p(x)$ share algorithmic information:
given $p(y|x)$, only  {\it specific} choices of $p(x)$ reproduce the Gaussian $p(y)$,
whereas generic choices of $p(x)$ would yield ``odd'' densities of the type on the right.}}
\end{figure}

To show that there are also joint distributions where $Y\rightarrow X$ is more natural, assume that $p(y)$ is Gaussian and the supports of $p(y|x=0)$ and $p(y|x=1)$ are $(-\infty,y_0]$ and $[y_0,\infty)$, respectively,
as shown in Fig.~\ref{Threshold}. 
One can easily think of a causal mechanism whose output $x$ is $1$  for all inputs $y$ above a certain threshold $y_0$, and $0$ otherwise.
Assuming $X\rightarrow Y$, we would require a mechanism that generates outputs $y$ from inputs $x$ according to $p(y|x)$. Given this mechanism,  there is only one distribution $p(x)$  of inputs
for which $p(y)$  is Gaussian. Hence,  the generation of the observed joint  distribution requires mutual adjustments of parameters for this causal model.  

In  Section~\ref{Just} we will describe more formal arguments 
that support this way of reasoning. It  is based on  ideas  in  \cite{LemeireD}  and \cite{Algorithmic}
to draw causal conclusions  not only from  {\it statistical} dependences. Instead,  also {\it  algorithmic} 
information can indicate causal  directions.

\section{Causal inference using second order exponential models}

\label{Sec:SO}

Here we
define a parametric family of Markov kernels  $p(x_j|pa_j)$ that  describe a simple way how $X_j$ is influenced by
its parents  $PA_j$. Without  loss of generality, we will only consider complete acyclic graphs, i.e.,
the parents of $X_j$ are given by $X_1,\dots,X_{j-1}$ (the general case is implicitly  included by setting
the corresponding parameters to zero).

The domains $T_j$ of  $X_j$ are subsets of $\R^{d_j}$ with integer Hausdorff dimension 
\cite{Federer} $\tilde{d}_j$,  i.e.,  we exclude fractal subsets.
In Sections~\ref{Ident} and \ref{Expe} we will consider, for instance, intervals in $\R$, circles in $\R^2$, and countable subsets
of $\R$.
We define
\begin{eqnarray}
p(x_1)&:=& \exp\left(\alpha^T_1 x_1 + x_1^T\beta_{11} x_1  - z_1\right) \nonumber \\   
p(x_j|x_1,\dots,x_{j-1})&:=&\exp\big( \alpha^T_j x_j +x^T_j\sum_{i\leq j} \beta_{ji} x_i -z_j(x_1,\dots,x_{j-1})\big) \label{SO}\,,
\end{eqnarray}
with vector-valued parameters $\alpha_j$ and matrix-valued parameters $\beta_{ji}$. 
The log-partition functions $z_j$ are given by
\[
z(x_1,\dots,x_{j-1}):=\log \int_{T_j} \exp\big( \alpha^T_j x_j +x^T_j\sum_{i\leq j} \beta_{ji} x_i \big)    d\mu_j(x_j)\,,
\]
where the reference measure $\mu_j$ is given by the product of the  Hausdorff measures \cite{Federer} of the corresponding dimensions $\tilde{d}_j$, and only parameters are allowed that yield normalizable densities. 
The term ``Hausdorff measure'' only formalizes the natural intuition of a volume of sufficiently 
well-behaved subsets of $\R^n$:
For a circle, for instance, it is given by the arc length, for countable  subsets  it is just the counting measure.

For every reordering $\pi$  of variables, the second order conditionals 
define a family of joint distributions $\cP_\pi$. The key observation on which our method
relies is that $\cP_\pi$ and  $\cP_{\pi'}$ need not to coincide if some of the variables $X_j$  have domains $T_j$ that are {\it proper} subsets of $\R^n$
(if, for instance, all $X_j$ can attain all values in $\R$, then $P_\pi$ is the  set of all non-degenerate  $n$-variate Gaussians for all $\pi$ and we cannot give 
preference to any causal ordering). 

Our inference rule reads: if there are causal  orders $\pi$ for which the observed density $p$ is  in $\cP_\pi$, prefer them to orderings $\tilde{\pi}$
for which  $p\not\in \cP_{\tilde{\pi}}$.
To apply this idea to finite  data  where $p$ is not available, we prefer the orderings $\pi$ for  which the Kullback-Leibler distance 
between  the empirical distribution and $p$ is minimized, i.e., the likelihood of the data is maximized.

In Section~\ref{Expe}, we discuss experiments with just two variables  $X,Y$. We have several cases where $X$ is binary and  $Y$ real-valued, and one example where 
$X$ is two-dimensional and attains values on a circle and $Y$ is real-valued. This shows that also causal structures containing  only continuous variables
can be dealt with by our method when some of the domains are restricted.
We now describe the algorithm.

\vspace{0.5cm}
\noindent
{\bf Second order model causal inference}

\begin{enumerate}

\item Given an $m\times n$ matrix of observations $x^{(i)}_j$.

\item Let  $X_1,\dots,X_n$ be an ordering $\pi$ of the variables. 

\item In the $j$th step, compute
$p(x_j|x_1,\dots,x_{j-1})$ 
 by minimizing the  conditional inverse log-likelihood
\[
L^\pi_j(\alpha,\beta):=-\alpha^T_j x_j -x^T_j\sum_{i\leq j} \beta_{ji} x_i +z(x_1,\dots,x_{j-1})\,,
\]
with 
\[
z_j(x_1,\dots,x_j):=-\log \int_{T_j} \exp(\alpha^T x_j +x^T_j \sum_{i\leq j}  \beta_{ji} x_i) d\mu_j(x_1)\,.
\]
To compute the partition function numerically,  we discretize and bound the domain to a finite set of points.

\item Compute the corresponding joint density $p_\pi (x_1,\dots,x_n)$ and its total log-likelihood 
\[
L^\pi:=\sum_{j=1}^n L^\pi_j
\]

\item  Select the causal orderings for which $L^\pi$ is minimal. This can be a unique ordering or a set of orderings
because not all orderings induce different families of joint distributions and because
values $L^\pi$ and $L^{\pi'}$ are considered equal if their 
 difference is below a certain threshold.

\end{enumerate}

For a preliminary justification  of the approach, we 
recall that
conditionals of this kind occur from maximizing the conditional Shannon entropy $S(X_j|X_1,\dots,X_{j-1})$ subject
to $P(X_1,\dots,X_{j-1})$ and subject to the
given first and second moments \cite{FrieGupt:2006}, for more details see also \cite{OccamsRazor}: 
\begin{eqnarray}
E(X_j)&=&c_j \\
E(X^T_j  X_i)&=& d_{ij}\,,
\end{eqnarray}
where $E(Z)$ denotes the expected value of a variable $Z$. 
For multi-dimensional $X_j$, 
the $c_j$ are  vectors and the $d_{ij}$  are  matrices. 
Bilinear constraints are the simplest constraints for which the entropy maximization yields interactions between
the variables $X_j$ (apart from this, linear constraints would not yield normalizable densities for unbounded domains).
In this sense, second order models generate the simplest non-trivial family of  conditional densities within a hierarchy of exponential models
\cite{Amari2001} that are given by entropy maximization subject to higher order moments.

\cite{OccamsRazor} provides a thermodynamic justification of second order models. 
The paper describes models of interacting physical systems, where  
the joint distribution is given by first maximizing the entropy of the {\it cause}-system
and then the conditional entropy of the {\it  effect}-system, given the distribution of the cause.
Both entropy maximizations are subject to energy constraints. If we assume that 
the physical energy is a polynomial of second order in
the relevant observables (which is not unusual in physics), we 
obtain exactly the second  order models introduced here.

\section{Identifiability results for special cases}

Here we describe examples that show how the restriction of the domains to proper subsets of $\R$ 
can make the models identifiable. A case with vector-valued variables  has already been described in
 \cite{SunLauderdale}, where we have considered the causal relation between 
the day in the year and the average temperature of the day.
The former takes values on a circle  
in $\R^2$, the latter is real-valued. Second order models
 from  day to temperature induce seasonal oscillations 
of the average temperature according to a sine function, which was closer to the truth
than the second order model from temperature to day in the year.

However, in the following examples we will restrict the  attention to 
one-dimensional variables.

\label{Ident}

\subsection{One binary and one real-valued variable}

\label{BiCo}

A simple case where cause and effect is identifiable in our model class is already given by the motivating example 
with a binary  variable $X$ and a variable $Y$ that can attain all values in $\R$.

\subsection*{Second order  model for $X\rightarrow Y$} 

Using both equations~(\ref{SO}),  we obtain
\begin{equation}\label{GaussMixPara}
p(x=1)=\gamma\quad \quad p(y|x=j)= \frac{1}{\sqrt{2\pi} \rho} e^{-\frac{(y-\nu_{j})^2}{2\rho^2}}\,,
\end{equation}
with parameters $\gamma,\nu_0,\nu_1,\rho$.
Both distributions $p(y|x=j)$ for $j=0,1$ are obviously  Gaussians with equal width and  different mean, i.e., $p(y)$ is a mixture of  two Gaussians (see Fig.~\ref{bimodal}, left).

\subsection*{Second  order model for $Y\rightarrow  X$}  

\begin{figure}
\centerline{\includegraphics[scale=0.40]{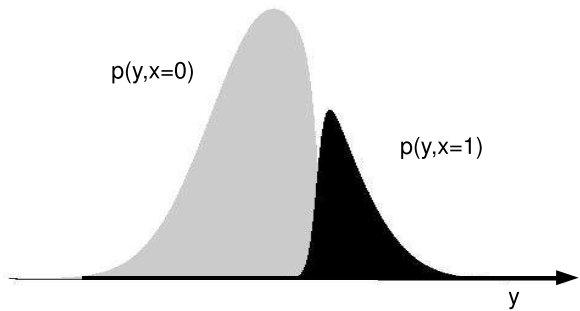}}
\caption{\label{Tanh}{\small Joint density $p(x,y)$ for binary $X$ and  real-valued $Y$ induced
by a second order model $Y\rightarrow X$. Here we  have chosen a relatively steep sigmoid function for $p(x=1|y)$, which leads to a steep decrease at the right of the left mode and the left of the right mode. An infinitely steep sigmoid function yields sharp thresholding as in Fig.~\ref{Threshold}.}}
\end{figure}

We obtain
\begin{equation}\label{SigmoidPara}
p(y)=\frac{1}{\sqrt{2\pi}\sigma} e^{-\frac{(y-\nu)^2}{2\sigma^2}}        \quad\quad p(x=1|y)=\frac{1}{2}\big(1+\tanh (\alpha y+\beta)\big)\,,
\end{equation}
with parameters  $\nu,\sigma,\alpha,\beta$,
where we have  used
\begin{equation}\label{ExpTanh}
\frac{e^a}{1+e^a}=\frac{1}{e^{-a}+1}=\frac{1}{2}(1+\tanh (a/2))\,.
\end{equation}
A typical joint distribution for the model $Y\rightarrow X$  is shown in Fig.~\ref{Tanh}.

Since mixtures of two different Gaussians can never yield a Gaussian as marginal distribution $p(y)$, the only joint distribution that is contained in the model classes for both directions  is
a product distribution of a Gaussian $p(y)$ and an arbitrary binary distribution $p(x)$.
This shows that the models are identifiable except for the trivial case  of independence.

Furthermore, eqs.~(\ref{GaussMixPara}) and (\ref{SigmoidPara}) show that our method is indeed  consistent with the
intuitive arguments we gave for the examples in the introduction: 
the Gaussian mixture in  Fig.~\ref{bimodal} is  a  second order model for  $X\rightarrow Y$ 
and the
example with  thresholding $y$ (Fig.~\ref{Threshold}) can be approximated by second order models 
for $Y\rightarrow  X$  via the limit $\alpha \to \infty$ in eq.~(\ref{SigmoidPara}).

\subsection{More than three binary variables}

\label{binary}
We first simplify equation~(\ref{SO}) for the case that all variables $X_1,\dots,X_n$  are binary.
Writing $an_j:=x_1,\dots,x_{j-1}$ for the ancestors of $x_j$,
we obtain
\[
p(x_j=1|an_j)=\frac{\exp (\alpha_j+\beta_{jj} + \sum_{i< j} \beta_{ji} x_i)}{1+\exp (\alpha_j+\beta_{jj} 
+ \sum_{i< j} \beta_{ji}x_i)} \,.
\]  
Using eq.~(\ref{ExpTanh}) yields
\begin{equation}\label{tanhRef}
p(x_j=1|an_j)=\frac{1}{2}\big(1+\tanh (\lambda_j +\sum_{i=1}^{j-1} \lambda_{ji} x_i)\big)\,.
\end{equation}
with
\[
\lambda_j:=\frac{1}{2}\left(\alpha_j+\beta_{jj}\right) \quad  \mbox{and} \quad \lambda_{ji}=\frac{1}{2}\beta_{ij} \quad \mbox{ for } \quad i=1,\dots,j-1 \,.
\]
  
The joint distributions induced by these conditionals do not coincide for all causal orders 
provided that $n\geq 4$. To show this, we  first
observe that second order models can approximate the causal relation  between
the inputs and the output of an $(n-1)$-bit OR gate. Then we show that the conditional probability for one
input, given the other $n-2$ inputs  and the output is significantly more complex than a second  order model
since
it requires polynomials of degree $n-2$ as argument of the $\tanh$-function 
(which corresponds to polynomials of degree $n-1$ in the exponent in the same way as second order models lead  to linear arguments
of $\tanh$).

The OR gate with input $X_1,\dots,X_{n-1}$ and output $X_n$ is described by
\[
p(x_n=1|an_n)=1-\prod_{i=1}^{n-1} (1-x_i)\,.
\]
Introducing a sequence of second order conditionals by
\[
p_k(x_n=1|an_n):=\frac{1}{2}\big( 1+\tanh (-k +2k \sum_{i=1}^{n-1} x_i )\big)\,,
\]
we have
\[
\lim_{k\to\infty} p_k(x_n|an_n)=p(x_n|an_n)\,,
\]
and thus they approximate the OR-gate.

Let the inputs  $X_1,\dots,X_{n-1}$ be sampled from the uniform distribution over $\{0,1\}^{n-1}$. 
We then have
\begin{equation}
p\left(x_1=1|x_2,x_3,\dots,x_{n-1},x_n=1\right)=
\label{CondB}
\left\{\begin{array}{cl} 
1 & \mbox{for} \, x_2=x_3=\cdots=x_{n-1}=0 \\
\frac{1}{2} & \mbox {otherwise} 
\end{array}\right. 
\end{equation}
and
\begin{equation}
\label{CondB2}
p\left(x_1=1|x_2= \cdots =x_n=0\right)= 0 \,.
\end{equation}
Note that the event $X_n=0$ and $X_i=1$ for some $i\in \left\{2,\ldots,n-1\right\}$ does not occur and the corresponding conditional probabilities need not to be specified.

We now show that the joint distribution cannot be approximated by second  order models if $X_n$ is not  the
last node. For symmetry reasons, it is sufficient to show that $p(x_1|x_2,\dots,x_n)$ has no second order model
approximation.  
If such an approximation existed, we would have
\begin{equation}\label{polyAppr}
p\left(x_1=1|x_2,\dots,x_{n}\right)=\lim_{k\to\infty}\frac{1}{2}\Big(1 +\tanh \big(q_k(x_2,\ldots,x_n\big))\Big)\,,
\end{equation}
where $q_k$ is a sequence of {\it linear} functions in $x_2,\dots,x_n$, see eq.~(\ref{tanhRef}). 
We prove that eq.~(\ref{polyAppr}) can indeed be satisfied with
$q_k$  of polynomials of order $n-2$, but not for any sequence of polynomials of lower order
(which shows that these non-causal conditional can be rather complex).
Introducing
\[
\tilde{q}_k(x_2,\dots,x_{n-1}):=q_k(x_2,\dots,x_{n-1},x_n=1)\,,
\]
eq.~(\ref{CondB}) is equivalent to
\begin{equation}\label{equivCondB}
\lim_{k\to\infty} \tilde{q}_k(x_2,\dots,x_{n-1},x_{n}=1)= \left\{\begin{array}{cl} 
\infty & \mbox{for} \, x_2=x_3=\cdots=x_{n-1}=0 \\
0 & \mbox {otherwise} 
\end{array}\right. 
\end{equation}
If the space  of polynomials of degree $n-3$ or lower contained such a  sequence $\tilde{q}_k$ , completeness of
finite dimensional real vector spaces  implies that it also contained
\[
g:=\lim_{k\to \infty} \frac{1}{\|\tilde{q}_k\|_1}\tilde{q}_k \,,
\]
which is given by
\[
g\left(x_2,\ldots,x_{n-1}\right)
=\left\{\begin{array}{cl} 
1 & \mbox{for} \,x_2=x_3=\cdots=x_{n-1}=0 \\
0 & \mbox{otherwise} \,.
\end{array}\right.
\]
However,
\[
g\left(x_2,\ldots,x_{n-1}\right)=\prod_{i=2}^{n-1} \left(1-x_i\right)\,,
\]
which is a polynomial of degree $n-2$. 
Hence, $\tilde{q}_k$ (and also $q_k$) consists at least of polynomials of order $n-2$.
To see that this bound is tight, set 
\[
q_k(x_2,\dots,x_n):=k\left( 2(x_n-1)-\prod_{i=2}^{n-1}(1-x_i)\right)\,,
\]
and observe that it satisfies eq.~(\ref{equivCondB}) and 
\[
\lim_{k\to\infty} q_k(x_2=0,\dots,x_n=0)=-\infty \,,
\]
and thus the corresponding conditionals satisfy asymptotically eqs.(\ref{CondB}) and (\ref{CondB2}).

By inverting logical values, the same proof applies to AND gates. Since AND and OR gates  are reasonable models
for many causal relations in real-life, it is remarkable that the corresponding non-causal  conditionals 
of the generated joint distribution already require exponential models of high order.  
Successful experiments with artificial and real-world data with four binary variables are briefly sketched in \cite{ESANNbinary}.

\section{Justification of our method by algorithmic information theory}

\label{Just}

\subsection{The principle of independent conditionals}

Since second order  models
provide a simple class of non-trivial conditional densities, Occam's Razor seems to strongly support the principle
of preferring the direction that admits such a model. However, Occam's Razor cannot justify why we should
try to find simple expressions for the causal conditional $P({\tt effect}|{\tt cause})$  instead of 
simple models for non-causal conditionals like $P({\tt cause}|{\tt effect})$.  
Here we present a justification that is based on recent algorithmic information theory based 
approaches to causal inference.
 
\cite{LemeireD} proposed to prefer those DAGs as {\it causal} 
hypotheses for which the shortest description
of the joint density $p(x_1,\dots,x_n)$ is given by separate descriptions
of causal conditionals $p(x_j|pa_j)$  in eq.~(\ref{Fac}).
 We will refer to this as  the principle 
of independent  conditionals (IC).
Here, the description length
is measured in terms of algorithmic information 
\cite{KolmoOr,Solomonoff,Chaitin},
sometimes also called ``Kolmogorov complexity''.
Even though it is hard to give a precise meaning
to this principle, it provides the leading motivation for our theory.

To show this, we reconsider one of the examples from the introduction.
We have argued
that the distribution in Fig.~\ref{Threshold} is unlikely to be generated by the causal structure $X\rightarrow Y$
because the observed distribution $p(x)$ 
is special among  all possible $\tilde{p}(x)$  since it is the only distribution  that yields
a Gaussian marginal $p(y)$ after feeding it into the conditional $p(y|x)$. Hence, 
after knowing  
$p(y|x)$, the input distribution $p(x)$  is simply described by ``the unique input that renders $p(y)$ 
Gaussian''. Thus, a description for $p(x,y)$  that  contains separate descriptions of $p(y|x)$ and $p(x)$  would contain redundant information and the  IC principle would be fail. 
We propose a slightly modified version of IC that will be more convenient to use because it refers to
algorithmic dependences between {\it un}conditional distributions:

\begin{Postulate}[independence of input and modified joint distr.]${}$\\ \label{modPost}
If the joint density $p(x,y)$ is generated by the causal structure $X\rightarrow Y$ then
the following condition must hold:

Let $\tilde{p}(x)$ be a  hypothetical input density that has been chosen without knowing 
$p(x,y)$. Define $\tilde{p}(x,y):= p(y|x) \tilde{p}(x)$.   
Then 
$p(x)$ and $\tilde{p}(x,y)$ are algorithmically independent. 
\end{Postulate}

The idea is that  $\tilde{p}(x,y)$ only contains algorithmic information about $p(y|x)$ and $\tilde{p}(x)$.  
The object $p(y|x)$ has been chosen independently of $p(x)$  ``by nature'', as in \cite{LemeireD}, and
$\tilde{p}(x)$ has been chosen independently of $p(y|x)$ by assumption.  

Due to the  lack of a precise meaning of  the concept of 
``algorithmic information of probability densities'', as 
it would be required
by \cite{LemeireD} and our modified postulate,  we will describe  arguments that avoid such concepts
but still rely on the above intuition.

\subsection{The framework for probability-free causal inference}

We therefore rephrase the probability-free approach to causal inference developed
by \cite{Algorithmic}. 
The idea is that causal inference in real life often does not rely on
statistical dependences. Instead, similarities between single objects indicate causal relations. Observing, for instance, 
that two carpets
contain the same  patterns makes us believe that designers have copied from each other (provided that the patterns are 
complex and not common).
\cite{Algorithmic}
develop a general framework for inferring causal graphs 
that connect individual objects  based upon {\it algorithmic} dependences.
Here, two objects are called algorithmically independent if their shortest joint description 
is given by the concatenations of their separate descriptions. It is assumed that every such description  
is a binary string $s$ formalizing all relevant properties of  an observation.
Then the Kolmogorov complexity $K(s)$ of $s$ is defined by the length of the shortest program that generates the 
output $s$ and then  stops. Conditional Kolmogorov complexity $K(s|t)$ is defined as the length of the shortest program that computes $s$ from the input $t$. If $t^*$ denotes the shortest
compression of $t$, $K(s|t^*)$ can be smaller than $K(s|t)$ because there  is no algorithmic way to obtain
the shortest compression (the difference between $K(s|t)$  and $K(s|t^*)$ can at most be logarithmic in the   
length of $t$ \cite{Vitanyi97}).

The strings $s$ and $t$  are conditionally independent, given $r$  if
\begin{equation}\label{Alin}
K(s,t|r)\approx K(s|r)+K(t|r)\,.
\end{equation}
As in the statistical setting, unconditional depedences indicate causal links between two objects $s$ and $t$: if
$
K(s,t)\ll K(s) +K(t)\,,
$
the descriptions can be better compressed jointly than independently and we postulate a causal connection. 
The following terminology \cite{GacsTromp} will be crucial:

\begin{Definition}[Algorithmic mutual information]${}$\\
For any two binary  strings  $s,t$,
the difference
\[
I(s:t):=K(s)+K(t)-K(s,t) \stackrel{+}{=}K(s)-K(s|t^*)\stackrel{+}{=} K(t)-K(t|s^*)
\]
is called the {\it algorithmic mutual} information between $s$ and  $t$. As usual in algorithmic information theory,
the symbol $\stackrel{+}{=}$ denotes equality up to a constant that is independent  of the strings $s,t$, but does depend on the Turing machine $K(.)$ refers to. 
\end{Definition}

To also infer causal directions 
we have postulated a causal Markov condition  stating conditional independence of every object from its non-effects, 
given its causes. We will here state an equivalent version (see Theorem~3 in \cite{Algorithmic}):

\begin{Postulate}[Algorithmic Markov condition]${}$\\ \label{algC}
Let $G$ be a DAG with the binary strings $s_1,\dots,s_n$ as nodes.
If every $s_j$ is the description of an object or an observation in real world and
$G$ formalizes the causal relation between them,  then 
the following condition must hold.

For any three sets $S,T,R \subset \{s_1,\dots,s_n\}$
we have
\[
S \independent T\,|R^*
\]
in the sense of eq.~(\ref{Alin}),
whenever $R$ d-separates $S$ and $T$ (for  the notion of d-separation see, e.g., \cite{Pearl2000}). 
Here we have slightly overloaded notation and identified 
the set of strings with their concatenation. 

Moreover, $R^*$ denotes the shortest compression of $R$.
In particular, we have 
\[
S\independent T\,,
\]
whenever $S$ and $T$ are d-separated (by the empty set).
Here, the threshold for counting dependences as significant is up to the 
decision of the researcher and not provided by the theory.
\end{Postulate}

\subsection{Distinguishing between cause and effect}

\begin{figure}
\centerline{
\includegraphics[scale=0.3]{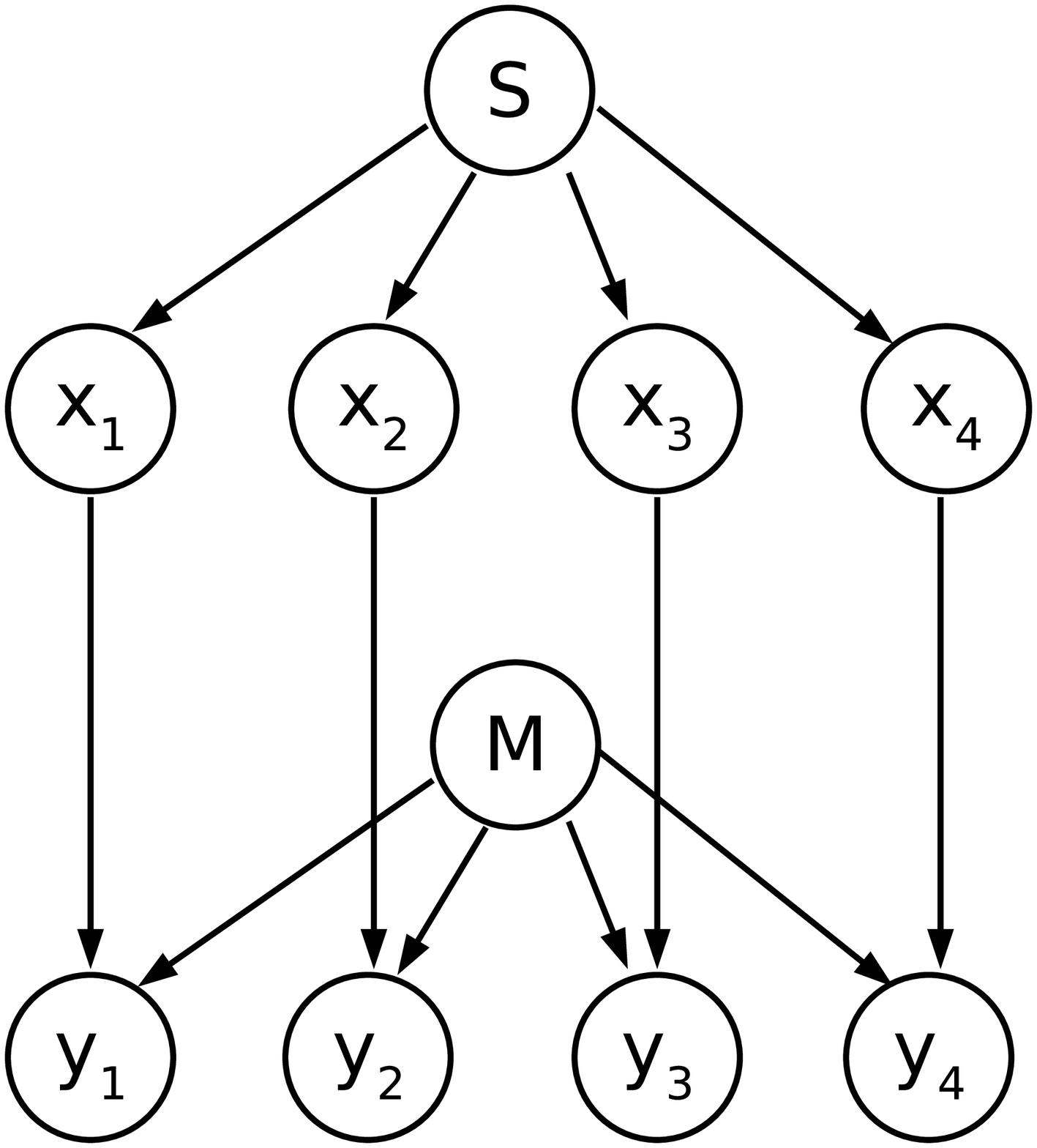}\hspace{1cm}\includegraphics[scale=0.3]{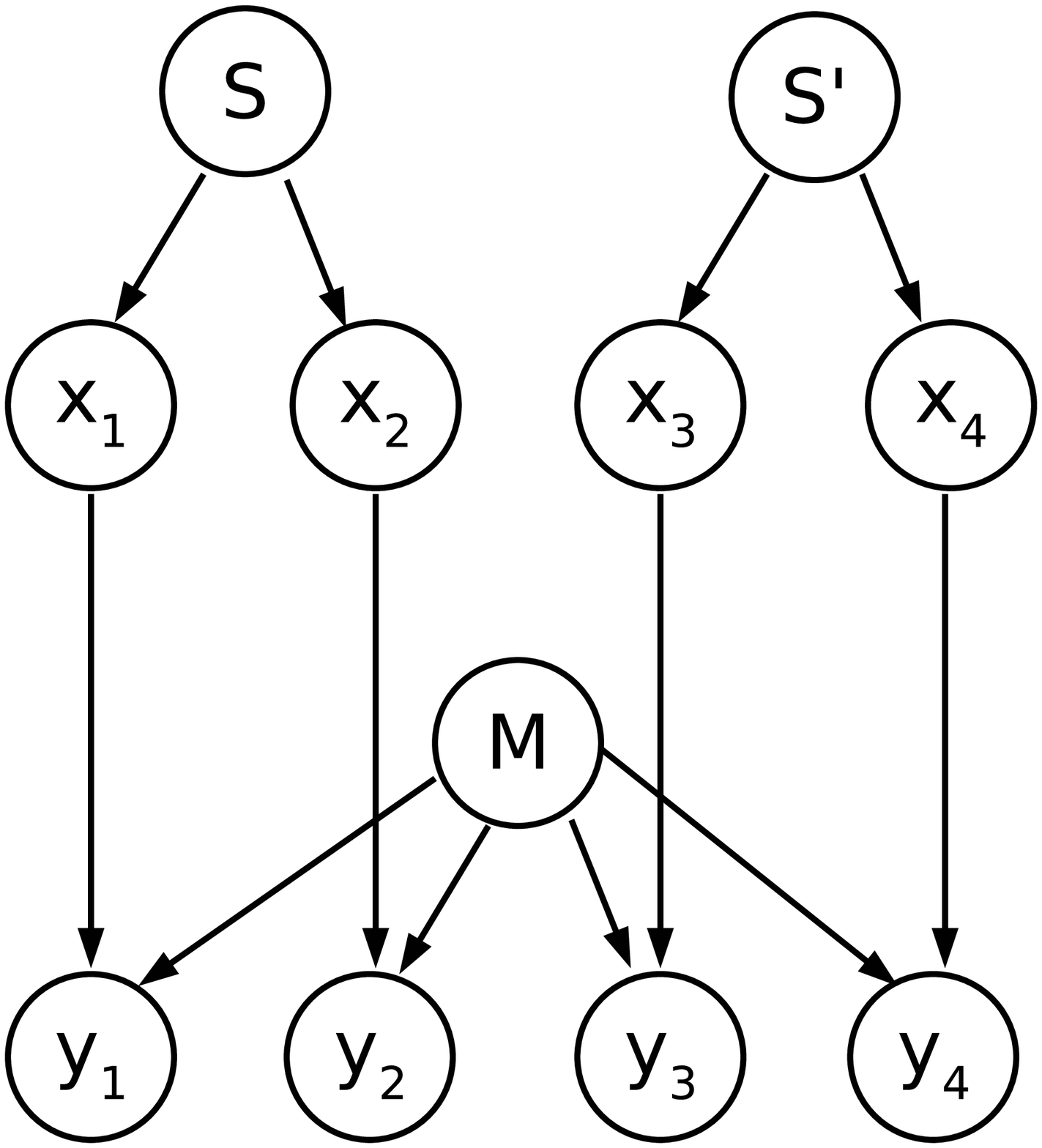}}
\caption{\label{Res}{\small Left: Causal structure obtained by resolving the statistical sample generated by the causal structure $X\rightarrow Y$ into single observations. Right:
Modified structure where the input comes from a different source $S'$ that samples according to a different distribution
$p'(x)$. Note that  $(x_1,x_2)$ and $(x_3,x_4,y_3,y_4)$ must be algorithmically independent because there is no unblocked path between these two sets of nodes.}}
\end{figure}

Based on the above framework and inspired by the IC-principle, 
\cite{Algorithmic} describe the following approach to  distinguishing between 
 $X\rightarrow Y$ and $Y\rightarrow X$  for two  random variables $X,Y$ after 
observing  the samples
 $(x_1,y_1),\dots,(x_k,y_k)$.  One considers the causal structure among $2k+2$ individual objects 
instead of a DAG with the two variables as nodes. 
These objects are: The $x$-values, the $y$-values, the source $S$ emitting $x$-values according to $p(x)$ and a machine
emitting $y$-values according to $p(y|x)$. The causal DAG connecting the objects is shown in  Fig.~\ref{Res}, left.
One may wonder why there are no arrows from the $x$-values to $M$ even though $M$ gets them as inputs.
The reason is that the object $M$ is not changed by the $x_j$, i.e., the conditional $p(y|x)$ remains constant.

It has been pointed out  \cite{Algorithmic} that the DAG   in Fig.~\ref{Res},  left, already 
imposes the algorithmic independence relation
\begin{equation}\label{dsep}
x_1,\dots,x_m \independent y_{m+1},\dots,y_k \, | (x_{m+1},\dots,x_k)^*  \quad \quad \forall m < k\,,
\end{equation}
and describes examples where this is violated after exchanging the role of  $X$ and $Y$.
This is an observable implication of the algorithmic independence of
the unobservable objects $S$ and $M$. The relevant information about $S$ and  $M$ is given  by 
$p(x)$ and $p(y|x)$, respectively, hence condition (\ref{dsep}) is closely linked to Lemeire's and  Dirkx's postulate.
\cite{Algorithmic} discusses toy examples for which the destinction between
$X\rightarrow Y$ and $Y\rightarrow X$ is  possible using condition~(\ref{dsep}).

For our purposes, it will be more convenient to work with  
a slightly  different condition that can be seen as a 
finite-sample counterpart of Postulate~\ref{modPost}.
To this end, we  consider Fig.~\ref{Res}, right. Let ${\bf  x}^1:= x_1,\dots,x_m$
be the sample of $x$-values from source $S$ (here $m=2$). ${\bf x}^2:=x_{m+1},\dots,x_k$ denote the
$x$-values from source $S'$ and ${\bf y}^2$ the corresponding $y$-values.
The d-separation criterion yields the unconditional relation
\begin{equation}\label{undsep}
{\bf x}^1 \independent  {\bf x}^2, {\bf y}^2\,. 
\end{equation}
Of course, we do not assume that we have the option to really  change the input distribution from $p(x)$  to $\tilde{p}(x)$ (i.e., replacing  the source $S$ with $S'$), otherwise
we  could  directly test whether  $X$ causes $Y$ by observing whether such an intervention also changes $p(y)$.
Our way of reasoning will be indirect: given that the true causal structure  is $X\rightarrow  Y$,   
we could simulate the effect of the intervention by choosing a subsample of $x$-values that is distributed
according to $\tilde{p}(x)$ and know that the  corresponding pairs  $(x,y)$ are distributed according 
to $p(y|x) \tilde{p}(x)$.  

\subsection{Applying the theory to second order models}

We now describe the violation of condition~(\ref{undsep}) for 
the example in Fig.~\ref{Threshold}. 
The true  model $Y\rightarrow X$ involves the parameters $\nu$, $\sigma$, and $\beta$ 
(mean and standard deviation of the Gaussian and threshold of $y$-values for which $x=1$).
We denote the corresponding density therefore by $p_{\nu,\sigma,\beta}$.
Now we consider the {\it  non-causal}  conditional $p_{\nu,\sigma,\beta}(y|x)$. 
One checks easily that different triples $(\nu,\sigma,\beta)$ indeed induce different
$p(y|x)$. On  the other  hand, there  is a unique input probability $p_\gamma (x=1)=\gamma$ such that the marginal
\[
p_{\gamma,\nu,\sigma,\beta}(y):= \sum_x p_\gamma (x) p_{\nu,\sigma,\beta}(y|x)
\]
is Gaussian:
\begin{equation}\label{fmu}
\gamma= f(\nu,\sigma,\beta):=\frac{1}{\sqrt{2\pi}\sigma} \int^\beta_{-\infty} e^{-\frac{(y-\nu)^2}{2\sigma^2}} dy \,.
\end{equation}
The fact that $f$ that does not  involve 
any free parameters
would  already  be in contradiction with Lemeire's
and Dirkx's postulate if $X\rightarrow Y$ were  true because $f$ only  has constant description length
and the parameters can be described with arbitrary accuracy. For sufficiently large accuracy, the description length of $\gamma$ thus
exceeds the length of $f$ and eq.~(\ref{fmu}) provides a shorter description for $\gamma$ than its explicit binary
representation. 

According to our finite-sample point of view, the 
parameters  $\gamma,\nu,\sigma,\beta$ must only be described up to an accuracy that corresponds to the error
made when estimating them from a finite sample: 
Observing ${\bf x}^1$, i.e., an ensemble of $x$-values drawn from $p(x)$, we can estimate $\gamma$ up to a certain accuracy. Similarly, we estimate $\nu,\sigma,\beta$ after observing ${\bf x}^2,{\bf  y}^2$
up to a certain accuracy. Then, the estimator for $\mu$ and the estimator  of the other parameters 
will approximately satisfy the functional relation (\ref{fmu}). This shows that 
${\bf  x}^1$, on the one hand, and ${\bf x}^2,{\bf y}^2$ on the other hand, share at least that amount
of algorithmic information shared by the estimators because the latter    
have only processed the information contained in the observations.  

For  general second order models, the argument reads as follows.
For $Y\rightarrow X$ we have  parameter vectors $\alpha_Y$ and $\alpha_{X|Y}$ 
for $p(y)$ and $p(x|y)$, respectively. The joint distribution is determined by $\alpha:=(\alpha_Y,\alpha_{X|Y})$.
Factorizing $p_\alpha(x,y)$ into the non-causal conditionals $p(x)$ and $p(y|x)$ leads 
to families $p_\theta(x)$ and $p_\eta(y|x)$, where $p_\theta(x)p_\eta(y|x)$ only is an element 
of the family $p_\alpha (x,y)$ if $\theta$ and $\eta$ satisfy a certain functional relation and thus share algorithmic information.
To translate this into algorithmic dependences between (real and hypothetical) observations, we feed  $p_\eta(y|x)$  with a modified input distribution
$p_{\theta'}(x)$ and observe that the generated $(x,y)$-pairs still share algorithmic information with those 
$x$-values that were 
sampled from the original input distribution because $\eta$ can be estimated from the new $(x,y)$ pairs 
and $\theta$ from the original $x$-values. If the parameters $\theta'$ and $\theta$ are algorithmically independent, the sources $S$ and $S'$ in Fig.~\ref{Res}, right, are independent and the causal hypothesis $X\rightarrow Y$ implies independence of ${\bf x}^1$ and $({\bf x}^2,{\bf y}^2)$. 

This way of reasoning can further be generalized as follows: assume that $p(x,y)$ can be described by 
$p_\theta(x)p_\eta(y|x)$ where $p_\theta(x)$ and $p_\eta(y|x)$ are some 
families of densities for which the map from $\theta,\eta,x,y$ to the corresponding 
probabilities is  a computable function.
Then $X\rightarrow Y$ can be rejected whenever one of the parameter is determined by the other one
via a computable function $f$ provided that the Kolmogorov complexity of both parameters is infinite (which can, of course, never be proved). 
The accuracy of estimating $\theta,\eta$ depends on the statistical distinguishability of 
the (conditional) densities from those for slightly modified $\theta+\Delta \theta,\eta+\Delta \eta$.
Therefore, Fisher information  of parametric families 
plays a crucial role in the following quantitative result:

\begin{Theorem}[dependent  parameters violate the algorithmic MC]${}$\\\label{MainTh}
Let $p_\theta(x)$ with $\theta\in \R^d$ and
$p_\eta(y|x)$ with 
 $\eta\in \R^{\tilde{d}}$ be computable families of continuously differentiable (conditional)  densities. 
Define the 
Fisher information matrix for $p_\theta(x)$ by
\[
(F_\theta)_{ij}:= \int\left(\frac{\partial \log p_\theta(x)}{\partial \theta_i}\frac{\partial \log p_\theta(x)}{\partial \theta_j}\right) p_\theta (x) d\mu_X(x)\,, 
\]
where $\mu_X$ defines the Hausdorff measure corresponding  to $X$.
Define the 
conditional Fisher information matrix for $p_\eta(y|x)$ with respect to the reference input distribution $p_\theta(x)$ by
\[
(\tilde{G}_{\eta,\theta})_{ij}:=\int \left(\frac{\partial \log p_\eta(y|x)}{\partial \eta_i}\frac{\partial \log p_\eta(y|x)}{\partial \eta_j}\right) p_\eta(y|x) p_\theta(x) d\mu_X(x) d\mu_Y(y)\,.
\]

Let $x_1,\dots,x_k$ be drawn from $p_\theta(x)$  
and $(x_{k+1},y_{k+1}),$$\dots,$$(x_{2k},y_{2k})$ from $p_{\theta'}(x)p_\eta (y|x)$
where 
$F_\theta$ and $G_{\eta,\theta'}$ are non-singular
and $\theta$ and $\eta$  are generic in the sense that
a description up to an error $\epsilon$ (in vector norm) requires
$d\log_2 \epsilon$ or $\tilde{d}\log_2 \epsilon$  bits, respectively.

Assume, moreover, that $\theta$ and $\eta$ are related as follows.  
If $d\geq \tilde{d}$, 
let $f(\theta)=\eta$ for some continuously differentiable function $f$ with $K(f)\Ceq 0$. For
$d<\tilde{d}$, let
$g(\eta)=\theta$ for some continuously differentiable $g$ with $K(g)\Ceq 0$.

Then the algorithmic mutual information between the
$x$-values sampled from the original distribution and the $(x,y)$-pairs generated by the modified input distribution
 satisfies asymptotically  almost surely
\[
I(x_1,\dots,x_k:x_{k+1},\dots,x_{2k},y_{k+1},\dots,y_{2k})\geq c \min\{d,\tilde{d}\}  \log k
\]
for every $c<1/2$.
\end{Theorem}

Note  that the requirement of ``generic'' parameter values (in the sense we used the term) 
can  be met by a model where  ``nature chooses'' 
them according to some prior density. Since the statement is only an asymptotic one, the theorem
holds regardless of the  prior. 

\vspace{0.5cm}
\noindent
Proof of Theorem~\ref{MainTh}: 
Assume first  that $d\geq \tilde{d}$.
We define an estimator $\hat{\theta}$ for $\theta$ by minimizing
\[
-\sum_{j=1}^k\log p_{\hat{\theta}}(x_j)\,.
\]
Hence, 
\[
\|\hat{\theta} -\theta\|<\frac{1}{(k \lambda_\theta)^c}
\]
with probability converging  to $1$ for $k\to\infty$ 
if $\lambda_\theta$ 
denotes the smallest eigenvalue of $F_\theta$. This is because $(\hat{\theta}-\theta)/\sqrt{k}$  is asymptotically a $d$-dimensional Gaussian with 
concentration matrix  $F_\theta$.
The standard deviation of the Gaussian is maximal for the direction corresponding to $\lambda_\theta$ and is 
then given by 
$1/\sqrt{\lambda_\theta}$.   

We construct an estimator $\hat{\eta}$ by minimizing the inverse loglikelihood
\[
-\sum_{j=k+1}^{2k}\log p_{\hat{\eta}} (y_j|x_j)\,.
\]
Since $G_{\eta,\theta'}$ is non-singular, $p_\eta(y|x)$ is a strict minimum of the expected loglikelihood.
As for the unconditional distributions above, $(\hat{\eta}-\eta)/\sqrt{k}$  is asymptotically Gaussian
and the probability for
\begin{equation}\label{etahat}
\|\hat{\eta}-\eta\|< \frac{1}{(k \nu_\eta)^c}
\end{equation}
tends to $1$ if $\nu_\eta$ denotes the smallest eigenvalue 
of $G_{\eta,\theta'}$.

Denoting the operator norm of the Jacobi matrix $Df(\theta)$ by $\|Df(\theta)\|$, we obtain  
\begin{equation}\label{fhat}
\|f(\hat{\theta})-f(\theta)\|\leq \|Df(\theta)\| \|\hat{\theta}-\theta\|+O(\|\hat{\theta}-\theta\|^2)\leq (\|Df(\theta)\|+\delta)
\|\hat{\theta}-\theta\|\,,
\end{equation}
where  the last inequality holds asymptotically almost surely
for any $\delta>0$. 

Due to the error bounds (\ref{fhat}) and
(\ref{etahat}) we have
\[
e:=\|f(\hat{\theta})-\hat{\eta}\| \leq 
\frac{\|Df(\theta)\|+\delta}{(k \lambda_\theta)^c} + \frac{1}{(k \nu_\eta)^c}
\] 
asymptotically 
with probability $1-\epsilon$  for any desired $\epsilon>0$. Since $\eta$ is a generic value,
the amount of information required  to specify it up to an accuracy $e$ grows 
asymptotically with
$-\log_2 e$ (up to some  negligible constant). On the other hand, $\hat{\eta}$ and $f(\hat{\theta})$ 
share  at least this amount of information because they also coincide up to an accuracy $e$. 
Hence,
\[
I(f(\hat{\theta}):\hat{\eta}) \stackrel{+}{\geq} -\log_2 e\,.
\]
Asymptotically, $-\log_2 e$ grows with  $c \log k$.
Hence the mutual information between $\hat{\eta}$ and  $f(\hat{\theta})$ is 
asymptotically larger than  $-cd \log_2 k$ bits for every  $c<1/2$. Hence we have 
\[
I(x_1,\dots,x_k:x_{k+1},\dots,x_{2k},y_{k+1},\dots,y_{2k}) \stackrel{+}{\geq} I(f(\hat{\theta}):\hat{\eta}) \geq cd \log_2 k\,.
\]
The first inequality follows because 
\[
I(a:b)\stackrel{+}{\geq} I(\tilde{a}:\tilde{b})
\]
whenever $K(\tilde{a}|a)\Ceq K(\tilde{b}|b)\Ceq 0$ 
(cf. Theorem II.7 in  \cite{GacsTromp}). Here
\[
K(f(\hat{\theta})|x_1,\dots,x_k)\Ceq 0
\]
 because $f(\hat{\theta})$ is computed from the $k$ observed $x$-values
by the above estimation procedure and the application of $f$. 
Likewise, $\hat{\eta}$ is derived from the observed $(x,y)$-pairs. 

The case for $d<\tilde{d}$ is shown similarly. We estimate $\eta$ and $\theta$ and show that they share algorithmic information because
$\theta$ is a simple function of $\eta$.
$\Box$

\vspace{0.5cm}

Now we present our main theorem stating that 
second order models between one binary and one real-valued variables induce joint distributions
whose {\it non-causal} marginals and conditionals are algorithmically dependent in the sense of Theorem~\ref{MainTh}: 

\begin{Theorem}[Justification of second order model inference]${}$\\\label{JustSO}
Let $X$ be a binary variable and $Y$ real-valued and 
the density of $p(x,y)$ be given by a second order model from
$Y$ to $X$ for some generic values of the parameters $\nu,\sigma,\alpha,\beta$ in eq.~(\ref{SigmoidPara}), left and right.
Then
 the causal hypothesis $X\rightarrow  Y$ contradicts the algorithmic Markov condition.
This is because the $x$-values sampled from $p(x)$ contain algorithmic information about the 
$(x,y)$-pairs obtained after changing the ``input'' distribution $p(x)$ (see Fig~\ref{Res}, right) and keeping $p(y|x)$.

Likewise, if
$p(x,y)$ admits a second order model from $X$ to $Y$ with generic values
$\gamma,\nu_0,\nu_1,\rho$ (see eq.~(\ref{GaussMixPara}), then $Y\rightarrow X$ 
must be rejected.

The amount of the shared algorithmic information grows at least logarithmically in the sample size.
\end{Theorem}

The remainder of this section is devoted to the proof of Theorem~\ref{JustSO} and a Lemma that is required for this purpose.
To show that the conditions of Theorem~\ref{MainTh} are met, we determine the parameter vectors $\theta,\eta$ of the non-causal conditionals,
show that they satisfy a functional relation and
that the Fisher information matrices are nonsingular. To prove the latter statement, we will use the following result:

\begin{Lemma}\label{central}
Let $p_\theta({\bf x})$ for all $\theta \in I\subset \R^d$ 
be a differentiable family of continuous positive definite densities on a probability space 
$\Omega  \subset \R^m$
with respect to the reference measure $\mu$. 
Assume there are $d$ points ${\bf x}_1,$ ${\bf x}_2$, $\dots,$ ${\bf x}_d$ such that the matrix $A(\theta)$ defined by
\[
A(\theta):=\left(\nabla p_\theta ({\bf x}_1),\dots,\nabla p_\theta ({\bf x}_d)\right)\,,
\]
or the matrix 
\[
\tilde{A}(\theta):=\left(\nabla \log p_\theta ({\bf x}_1),\dots,\nabla \log p_\theta ({\bf x}_d)\right)
\]
is non-singular. Then the Fisher information matrix
$F_\theta$ is non-singular.
\end{Lemma}

\noindent
Proof: the Fisher information matrix can be rewritten as
\[
(F_\theta)_{ij}=\int_\Omega \frac{1}{p_\theta({\bf x})} \frac{\partial p_\theta  ({\bf x})}{\partial \theta_i} 
\frac{\partial p_\theta  ({\bf x})}{\partial \theta_j} d\mu({\bf x})\,.
\]
Hence,
\[
F_\theta=\int_\Omega \frac{1}{p_\theta({\bf x})} (\nabla p_\theta ({\bf x})) (\nabla p_\theta ({\bf x}))^T d\mu({\bf x})=\int_\Omega 
p_\theta({\bf x})
(\nabla \log p_\theta ({\bf x}))(\nabla \log p_\theta ({\bf x}))^T d\mu({\bf x})\,.
\]
It thus is the weighted integral over all rank one matrices
\[
(\nabla p_\theta ({\bf x})) (\nabla p_\theta({\bf x}))^T\,.
\]
At the same time, it can also be written as 
a weighted integral over all
\[
(\nabla \log p_\theta ({\bf x})) (\nabla \log p_\theta({\bf x}))^T\,.
\]
Note that for any vector-valued continuous function  $v$ and strictly positive scalar function 
$q$, the image of the matrix
\[
\int q({\bf x}) v({\bf x})v({\bf x})^T d\mu({\bf x})
\]
is given by the span of all $v({\bf x})$. 
$F_\theta$ thus is the span over all $\{\nabla p_\theta  ({\bf x})\}_{{\bf x}}$
and, at the same time, the span over all $\{\nabla \log p_\theta ({\bf x})\}_{{\bf x}}$.
$\Box$

\vspace{0.5cm}
\noindent
We are now able to prove the main theorem:

\vspace{0.3cm}
\noindent
Proof (of Theorem~\ref{JustSO}): 
First consider the case where $p(x,y)$ has a  second order model from $Y$ to $X$. 
To apply  Theorem~\ref{MainTh} we have to show that
$G_{\theta,\eta}$ is non-singular.
We can use Lemma~\ref{central} even though it  is not explicitly stated for {\it conditional} densities 
 because we can apply the latter  
to the joint density $p_\eta({\bf x}):=p_\theta(x)p_\eta(y|x)$ for ${\bf  x}:=(x,y)$ and fixed $\theta$.
Then 
\[
\nabla \log p_\eta  (x,y)  = \nabla \log p_\eta (y|x)\,,
\] 
and
\[
\nabla p_\eta (x,y)= p_\theta (x) \,\nabla p_\eta  (y|x)\,,
\]
i.e., it is sufficient to check whether the gradients of the {\it conditional} or its logarithm 
span a $\tilde{d}$-dimensional space. 
We have 
\[
p_{\sigma,\nu,\alpha,\beta}(x=1,y)=
\frac{1}{2\sigma \sqrt{2 \pi}}  \left(1+\tanh (\alpha y+\beta)\right) e^{-\frac{(y-\nu)^2}{2\sigma^2}} =
\frac{e^{-\frac{(y-\nu)^2}{2\sigma^2}}}{\sigma  \sqrt{2 \pi}(1+e^{2\alpha y+2\beta})}\,,
\]
where we have
used eq.~(\ref{ExpTanh}).
This yields
\begin{equation}\label{x=1}
p_{\sigma,\nu,\alpha,\beta}(x=1)= \frac{1}{\sigma  \sqrt{2 \pi}}\int \frac{e^{-\frac{(y-\nu)^2}{2\sigma^2}}}{1+e^{2\alpha y+2\beta}} dy\,.
\end{equation}
Introducing the parameter vector $\eta:=(\sigma,\nu,\alpha,\beta)$ we obtain
\[
p_{\eta}(y|x=1)= \frac{1}{p_\eta (x=1)} \frac{e^{-\frac{(y-\nu)^2}{2\sigma^2}}}{\sigma \sqrt{2\pi}(1+e^{2\alpha y+2\beta})}\,,
\]
where the input distribution $p(x)$ still is formally parameterized by $\eta$ and will be written in terms of one relevant parameter $\theta$ below. 
In the appendix we provide $4$ points $y_1,\dots,y_4$ and a value $\eta=\eta_0$ for which the vectors $\nabla p_\eta(y_j|x=1)$ are linearly independent. Hence   $G_{\eta,\theta}$ is non-singular for $\eta_0$ and all $\theta$.
All entries of $G_{\theta,\eta}$ are analytical functions in every component of  $\eta$
because they are uniformly  converging integrals over analytical functions.
Hence, regularity  of  $G_{\eta,\theta}$ for one $\eta$ already shows 
regularity for {\it generic} $\eta$.  

Now we parameterize $p(x)$ by an  one-dimensional parameter 
\[
\theta:=p_\eta (x=1)=g(\eta)\,,
\]
where $p_\eta(x)$ is given by the integral in eq.~(\ref{x=1}).  This defines the family of densities $p_\theta (x)$ 
via
\[
p_\theta (x=1):=\theta\,.
\]
Hence $F_\theta$ is one-dimensional. It  is clearly non-singular for generic  $\theta$ because 
\[
\frac{\partial p_\theta (x=1)}{\partial \theta} \neq 0\,.
\]
Using $K(g)\Ceq 0$,  Theorem~\ref{MainTh} 
shows  that the $x$ values sampled from $p_\theta(x)$ share algorithmic information with 
the  $(x,y)$-pairs sampled  from $p_{\theta'}(x)p_\eta(y|x)$.

Now consider the case that there is  a  second order model from $X$ to $Y$. 
Hence
\[
p_{\theta}(y)= \frac{1}{\rho \sqrt{2\pi}} \left((1-\gamma)e^{-\frac{(y-\nu_0)^2}{2  \rho^2}}  +   
\gamma e^{-\frac{(y-\nu_1)^2}{2  \rho^2}} \right)\,,
\]
with the parameter vector $\theta=(\gamma,\nu_0,\nu_1,\rho)$.
Note that we now apply Theorem~\ref{MainTh} with exchanging the role of $X$ and $Y$.
To show that $F_\theta$ is non-singular we 
compute $\nabla p_\theta (y)$  and find points $y_1,\dots,y_4$ and a value $\theta$ such that
the corresponding gradients are linearly independent (see Appendix~\ref{Kram}). Hence $F_{\theta}$ 
is  nonsingular due to Lemma~\ref{central}. As above, this also holds for generic $\theta$.

For the conditional density of $X$ given $Y$,  only a function of $\theta$ is relevant (as above) 
but we start by writing it first 
in terms of $\theta$ and reduce the parameter space later to the relevant part:
\begin{eqnarray*}
p_{\gamma,\nu_0,\nu_1,\rho} (x=1|y)&=&
\gamma e^{ -\frac{ (y-\nu_1)^2 }{ 2\rho^2 } } \left[(1-\gamma) e^{-\frac{(y-\nu_0)^2}{2\rho^2}} +\gamma e^{- \frac{(y-\nu_1)^2}{2\rho^2} }  \right]^{-1}\,.
\end{eqnarray*}  
Introducing
\begin{equation}\label{al}
\alpha:=\frac{1}{\rho^2}(\nu_0-\nu_1)
\end{equation}
and
\begin{equation}\label{bet}
\beta:=\frac{1-\gamma}{\gamma} \exp\left(\frac{\nu^2_1-\nu^2_0}{2\rho^2}\right)\,,
\end{equation}
the conditional is of  the form
\[
p_{\gamma,\alpha,\beta}(x=1|y) =\frac{1}{\beta e^{\alpha y} +1}   
\,.
\]
We
define $\eta:=(\alpha,\beta)$  and check that $G_{\eta,\theta}$ is non-singular.
For doing so, we compute 
$\nabla p_\eta (x=1|y)$ and find values $\eta_0$ and $y_1,\dots,y_2,y_3$ such that the gradients are linearly independent (Appendix). Hence $G_{\eta,\theta}$ is non-singular for one $\eta_0$ and all $\theta$ 
and thus also for generic pairs $\eta,\theta$.
The function $g$ is given by
$
g(\gamma,\nu_0,\nu_1,\rho):=(\alpha,\beta)$ with $\alpha$ and  $\beta$ as in eqs.~(\ref{al}) and (\ref{bet}),
which satisfies $K(g)\Ceq 0$. This shows that the $y$-values sampled from $p_{\theta}(y)$ share  
algorithmic information with the $(x,y)$-pairs sampled  from $p_{\theta'}(y)p_\eta(x|y)$  by Theorem~\ref{MainTh}.
$\Box$

\section{Experiments}

\label{Expe}

We conducted 8 experiments with real-world data for which the causal structure is known. In  all cases we had pairs  of variables
where one is the cause and one the effect. Even though there may also be hidden common causes, prior knowledge 
strongly suggests that a significant part  of the dependences are due to an arrow from one variable to the other.  
The selection of datasets was based on the following criteria:
We have chosen several examples where 
one variable is binary and  the other one is either continuous or 
discrete with a  wide range, because this is the case where identifiability becomes most obvious (see Subsection~\ref{BiCo}).
To demonstrate that we have identifiability for various types of value  sets we have also included an  example
with a variable of angular-type and example with positive variables. 
The restriction to positive values,  however, only leads to significantly different distributions
for different causal directions  if there is enough probability close  to the boundary. Otherwise, the second order models yield almost bivariate Gaussians and the direction is not identifiable. Most examples of the data base 
``cause effect pairs'' in the NIPS 2008 causality competition \cite{MooijJanzingSchoelkopf08} are of this type, except for  the examples with ``altitude''.  

Our algorithm constructs the domains by binning 
the observed values into intervals of equal length instead of asking for the range  as additional input.
If the differences of  the loglikelihoods are too small, our algorithm will not decide for either of the
causal directions. We have set the treshold to 
\[
|L_\rightarrow -L_\leftarrow |\leq \frac{1}{10000}\frac{L_\rightarrow +L_\leftarrow}{2} \,.
\]
The choice  of this threshold, however, is the result  of our limited number of experiments.
Our theory in Section~\ref{Just} only states the following: if the  true distribution  {\it perfectly} coincides
with a second order model in one direction but not the other, the latter one  has  to be  rejected because
this causal structure would require unlikely adjustments. For  the case where the distribution is only close to a
second order model it is hard to analyze how close it should be to justify our causal conclusion.
The answer to this question is left  to the future.

\subsection*{Meteorological data}

Experiment No.~1 
considers the altitude and average temperature of 675 locations in Germany \cite{Wetter}.  
The statistical dependence between both variables 
is  very obvious from scatter plots and one observes an almost linear decrease of the temperature with increasing altitude.
The fact that a significant part of the  points are close to altitude $0$ (i.e., the minimal value) is important
for identifiability of the causal direction because the restriction of the domain to positive values can only be relevant in this case. 
 
Experiment No.~2 studies the relation between altitude and precipitation of 
4748 locations in Germany \cite{Wetter}. 
Here both variables are positive-valued,  which also leads  to  different models
in the two directions.

In experiment No.~3, we were given the daily temperature averages of 9162 consecutive days between 1979 and 2004 in Furtwangen, Germany \cite{Bernward}.
The seasonal cycle leads  to a strong statistical dependence between the variable
{\tt day in the year} (represented as a point on the unit circle $S^1\subset  \R^2$) and  {\tt temperature}, where the former
should be considered as  the cause since it describes the position of the earth on its orbit around the sun.

\subsection*{Human categorization} 

Our experiments No.~4 and No.~5  consider
two datasets from the same psychological experiment on human categorization. 
The subjects are shown artificially generated faces that interpolate between male and female faces \cite{Regine}.
The interpolation correponds to switching a parameter between $1$ and $15$ (in integer steps).
The subjects are asked to decide whether the face is male ({\tt answer}=0)  or female  ({\tt answer}=1).
The  experimentalist has chosen parameter values  according to a uniform distribution on
$\{1,\dots,15\}$. 

No.~4 studies the  relation between {\tt  parameter} and {\tt answer}.
Since the experimentalist chose uniform distribution over $\{1,\dots,15\}$ and the dependence of the probability for {\tt answer}$=1$ is  close to a  sigmoid function, the empirical distribution is here  very close to the second order model 
corresponding to the correct causal structure {\tt parameter} $\rightarrow$ {\tt answer}. 

Our experiment  No.~5
studies the relation between the response time and the parameter values.
Since the response  time is minimal for both extremes  in the parameter values, we have  strongly non-linear 
interactions that cannot be captured by second-order models. It  is  therefore not surprising 
that there is no decision in this case.

\subsection*{Census data}

Experiments No.~6 and No.~7  consider census data from 35.326 persons in the  USA \cite{MLRCensus}.
In No.~6, the relation between age and marital  status is studied. The latter takes  the  two values
$0$ for  never married and $1$ for married, divorced, or widowed.
No.~7 considers the relation between gender and  income. Here we assume that the gender is almost randomized
by nature and there we thus expect  no confounding to any observable variable.

\subsection*{Constituents of wine}

Experiment No.8 considers the concentration of proline in wine from two different cultivars. 
We assume  that the binary variable {\tt cultivar} is the cause, even though one cannot exclude that the proline level 
(if relevant for  the taste)
directly influenced the decision of the cultivar to choose this sort of wine.

\subsection*{List of results} 

The results are shown in the below table. The ground truth is always that variable 1  influences variable 2, i.e.,
we have one  wrong result and no decision in two cases.

\vspace{0.5cm}
\hspace{-1.5cm}
\begin{tabular}{lllllll}
\hline
No. & variable 1, domain &  variable 2,  value  set & $L_\rightarrow$ & $L_\leftarrow$ &  result \\
\hline
1&{\tt altitude}, $\R^+$  & {\tt temperature}, $\R$ & 3.3697 & 3.4366 &  $\rightarrow$ \\
2&{\tt altitude}, ${\R^+}$ & {\tt precipitation}, $\R^+$ & 3.5885 & 3.6343 & $\rightarrow $\\
3&{\tt day of the year}, $S^1$ & {\tt temperature}, $\R$ & 5.7448 &   5.7527 & $\rightarrow $ \\
4&{\tt parameter},\{1,\dots,15\}& {\tt answer},  $\{0,1\}$ & 4.1143 & 3.1150 & $\rightarrow$ \\
5&{\tt parameter},\{1,\dots,15 \}& {\tt time}, $\R^+$ & 3.9873 & 3.9873 & ? \\
6&{\tt age}, $\R^+$ & {\tt marrital status}, $\{0,1\}$ & 4.9918 &  4.9920 & $ ? $ \\
7&{\tt sex}, $\{0,1\}$ &{\tt  income}, $\R^+$ & 3.8770 & 3.8758  & $\leftarrow$ \\   
8&{\tt cultivar}, $\{0,1\}$ & {\tt proline},  $\R^+$ & 3.9209 & 3.9496 & $\rightarrow $  \\ 
\hline
\end{tabular}

\section{Discussion and relations to independence-based causal inference}

In section~\ref{Just} we have shown for a special case that
the model $X\rightarrow Y$ must be rejected if there is a second order model from $Y$ to $X$ because 
it required specific mutual adjustments of $p(x)$ and $p(y|x)$ to admit such a model.
We have already mentioned that this is the same idea as rejecting unfaithful distributions.
Indeed, \cite{LemeireD} argued that the Markov kernels in unfaithful distributions share algorithmic information. Hence 
algorithmic information theory provides a unifying framework for independence-based approaches and those that impose constraints on the shapes of conditional densities.

The following example makes this link even closer because it shows that 
in some situations the same constraints on a joint distribution
may appear as {\it independence} constraints from one 
point of view and as constraints on the {\it shape} of conditionals from an other perspective. 
Consider the causal chain
\begin{equation}\label{timeorder}
X_1\rightarrow X_2\rightarrow \cdots \rightarrow X_n\,,
\end{equation}
where every  $X_j$  is a vector of dimension $d$.
Structures of this kind occur, for instance, 
if $X_j$ represents the state  of some  system at time $t$ and the dynamics is generated by a first order Markov  process.
Due to the causal Markov condition the joint distribution factorizes into
\[
p(x_1)p(x_2|x_1)\cdots p(x_n|p_{n-1})\,,
\] 
but no constraints are  imposed on the conditionals   $p(x_j|x_{j-1})$.

\begin{figure}
\centerline{\includegraphics[scale=0.30]{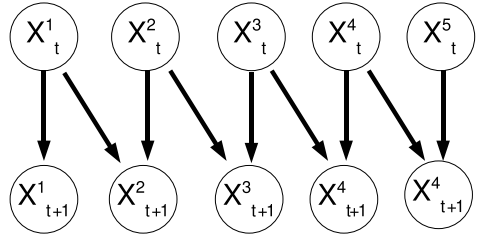}}
\caption{\label{Layers}{\small Two Layers in  the causal chain. If the components are  only influenced  by horizontally adjacent ones from the layer above, the Markov condition further simplifies the forward time conditional.}}
\end{figure}

Assume  now we consider each  component $X_j^{(i)}$ of layer $j$ as a variable in its  own right and thus obtain a causal structure
between $\tilde{n}:=n d$ variables. Assuming that no component $X_j^{(i)}$ is influenced by components of the same layer,
$p(x_j|x_{j-1})$ must be of the form
\begin{equation}\label{time_series}
p(x_j|x_{j-1})=\Pi_{i=1}^{d} p(x^{(i)}_j|x_{j-1})\,.
\end{equation}
Moreover, if we assume that every $X^{(i)}_j$ is only influenced by some of the variables in the previous layer,
the conditional further simplifies into
\[
p(x_j|x_{j-1})=\Pi_{i=1}^{d} p(x^{(i)}_j|pa_{ji})\,,
\]
where $pa_{ji}$ denote the values of $PA_{ij}$, i.e., the  parents of $X^{(i)}_j$ (Fig.~\ref{Layers}). 
Hence, the fine-structure of the causal graph imposes constraints on $p(x_j|x_{j-1})$ that are not imposed  by the coarse-grained structure. 

Assume we  are given data from the above time series,  but  it is not known whether the true causal structure reads
\[
X_n \rightarrow X_{n-1}\rightarrow \cdots \rightarrow X_1\,,
\]
or the one in
(\ref{timeorder}). When resolving the vectors in their  components, we have to reject the latter hypothesis because the independences 
\begin{equation}\label{sim}
X_j^{(i)} \independent X_j^{(i')} \, |X_{j-1}
\end{equation}
would violate  faithfulness. From the coarse-grained perspective, we can only reject the causal hypothesis by imposing an appropriate simplicity principle of conditionals.
Then we conclude that (\ref{timeorder}) is more likely to be true because the Markov kernels are simpler
because they satisfy  eq.~(\ref{sim}). 
Finding further useful simplicity constraints has  to be left to the future.

\section{Appendix}

\label{Kram}

\subsection{Matrix for $p(y|x)$}

We have:
\begin{eqnarray*}
\log p_{\sigma,\nu,\alpha,\beta}(y|x=1)&=& -\log \sigma -\log \sqrt{2 \pi} -\frac{(y-\nu)^2}{2\sigma^2} -\log\left(e^{2\alpha y+2\beta}+1\right)\\&&-\log p_{\sigma,\nu,\alpha,\beta}  (x=1)\,.  
\end{eqnarray*}
Hence
\[
\begin{array}{ccccc}
\frac{\partial \log p_\eta (y|x=1)}{\partial \nu}&=&  -\frac{y-\nu}{\sigma^2}- \frac{\partial \log  p_\eta (x=1)}{\partial \nu} &=:&h_1(y)\\
\frac{\partial \log p_\eta (y|x=1)}{\partial \sigma}&=&-\frac{1}{\sigma}+ \frac{(y-\nu)^2}{\sigma^3}- \frac{\partial \log  p_\eta (x=1)}{\partial \sigma}&=:&h_2(y)\\
\frac{\partial \log p_\eta (y|x=1)}{\partial \alpha}&=& - 2y \frac{e^{2\alpha  y +2\beta}}{e^{2\alpha y+2\beta}+1} - \frac{\partial \log  p_\eta (x=1)}{\partial \alpha}&=:&h_3(y)\\
\frac{\partial \log p_\eta (y|x=1)}{\partial \beta}&=& -2  \frac{ e^{2\alpha  y +2\beta}    }{e^{2\alpha y+2\beta}+1}- \frac{\partial \log  p_\eta (x=1)}{\partial \beta} &=:& h_4(y)\,,
\end{array}
\]
Intuitively, it is quite evident that the functions $h_j$ are linearly independent for generic $\eta$ because $h_1$  
contains linear terms  in  $y$, $h_2$ is  a polynomial of degree two in $y$, $h_3$ contains $y$ and an expression  with an exponential function in  the denominator, $h_3$ contains only the exponential expression in the denominator. We can thus find points $y_1,\dots,y_4$
such that the row vectors 
$(h_j(y_1),h_j(y_2),h_j(y_3),h_j(y_4))$ are linearly independent.
Instead of proving  this directly (which would involve  derivatives of the logarithms
of marginals), we use the  following indirect argument:
Choose $5$ values $y'_0,\dots,y'_4$ and consider the rows
\begin{equation}\label{5dim}
\left[h_j(y'_0),h_j(y'_1)\dots,h_j(y'_4)\right] \quad j=1,\dots,4\,.
\end{equation}
Consider the projection of 
$\R^5$ onto the  quotient space $\R^5/\R (1,1,1,1,1)$  and represent
the images of the vectors (\ref{5dim}) by
\[
\left[h_j(y'_1)-h_j(y'_0),h_j(y'_2)-h_j(y'_0),\dots,h_j(y'_4)-h_j(y'_0)\right] \quad j=1,\dots,4\,.
\]
Check that these $4$ vectors are linearly independent (which can fortunately 
be done without computing the derivatives of $\log_{\sigma,\nu,\alpha,\beta}  p(x=1)$), hence the 
row vectors (\ref{5dim}) are  independent, too. We can thus select $4$ values $y_1,\dots,y_4$ from $y_0',\dots,y_4'$
such that the rows $(h_j(y_1),\dots,h_j(y_4))$ are  independent.
We  have numerically checked this for $\nu=\sigma=\alpha=\beta=1,y'_j=j$.

\subsection{Matrix for $p(y)$}

The coefficients of $\nabla p_\theta (y)$ read:
\begin{eqnarray*}
\frac{\partial p_\theta(y)}{\partial \gamma} &=&  \frac{1}{\sqrt{ 2\pi}\rho} \left(-e^{-\frac{(y-\nu_0)^2}{2\rho^2}} + e^{-\frac{(y-\nu_1)^2}{2\rho^2}}\right) \\
\frac{\partial p_\theta(y)}{\partial \nu_0} &=&\frac{1}{\sqrt{ 2\pi}\rho} \frac{(1-\gamma)(y-\nu_0)}{\rho^2} e^{-\frac{(y-\nu_0)^2}{2\rho^2}} \\
\frac{\partial p_\theta(y)}{\partial \nu_1} &=& \frac{1}{\sqrt{ 2\pi}\rho} \frac{\gamma(y-\nu_1)}{\rho^2} e^{-\frac{(y-\nu_1)^2}{2\rho^2}}  \\
\frac{\partial p_\theta(y)}{\partial \rho} &=&\frac{1}{\sqrt{ 2\pi}\rho} \left\{  (1-\gamma)\left(\frac{(y-\nu_0)^2}{\rho^3}-\frac{1}{\rho}\right) e^{\frac{(y-\nu_0)^2}{2\rho^2}} \right. \\&&  \left.
\hspace{4.5cm}+
\gamma \left(\frac{(y-\nu_1)^2}{\rho^3}-\frac{1}{\rho}\right) e^{-\frac{(y-\nu_1)^2}{2\rho^2}}\right\}\,.
\end{eqnarray*}
The vectors are linearly independent for the points $y_j=j$ for $j=1,\dots,4$ with $\gamma=1/2,\nu_0=0,\nu_1=1,\rho=1$.

\subsection{Matrix for $p(x|y)$}

Introducing the   function $h$ with 
\[
h_y(\eta):=\beta e^{\alpha y} +1
\]
we have
\[
p_\eta (x=1|y)=1/h_y(\eta)
\]
and  thus 
\[
\nabla p_\eta  (x=1|y)=-\frac{1}{h_y^2(\eta)} \nabla h_y(\eta)\,,
\]
with
\[
\frac{\partial h_y (\eta)}{\partial \alpha}= y \beta e^{\alpha y}  \quad \hbox{ and }  \quad 
\frac{\partial h_y (\eta)}{\partial \beta}=   e^{\alpha y} \,.     
\]
Since the functions $h_1(y):=y \beta e^{\alpha y}$ and  $h_2(y):=e^{\alpha y}$ are linearly independent
for generic $\alpha,\beta$, we can obviously find values $y_1,y_2$ such that $\nabla p_\eta  (x=1|y_1)$
and $\nabla p_\eta  (x=1|y_2)$  are linearly independent.

\section*{Acknowledgements}

Thanks to Bastian Steudel and Jonas Peters for several comments on an earlier version.


\end{document}